# Modelling Social Structures and Hierarchies in Language Evolution

Martin Bachwerk and Carl Vogel

Computational Linguistics Group,
School of Computer Science and Statistics,
Trinity College, Dublin 2, Ireland
{bachwerm,vogel}@tcd.ie

**Abstract.** Language evolution might have preferred certain prior social configurations over others. Experiments conducted with models of different social structures (varying subgroup interactions and the role of a dominant interlocutor) suggest that having isolated agent groups rather than an interconnected agent is more advantageous for the emergence of a social communication system. Distinctive groups that are closely connected by communication yield systems less like natural language than fully isolated groups inhabiting the same world. Furthermore, the addition of a dominant male who is asymmetrically favoured as a hearer, and equally likely to be a speaker has no positive influence on the disjoint groups.

## 1 Introduction

The question of how human language could have emerged from an animal-like communication system is not only fascinating from an evolutionary point of view, but also has broad ramifications in the area of natural language and speech development. If we could understand how our extremely distant ancestors learned to associate meaning with seemingly arbitrary symbols, be those symbols gestures or sounds, then we should have an easier time of engineering artificial systems capable of comparable levels of intelligence.

Although speculation about the origin of human language has gone on for centuries, the problem has only relatively recently been scrutinized in empirically oriented disciplines, including anthropology and evolutionary biology [8], linguistics [13], artificial intelligence [20,6], and computer science [15,19]. The contribution of the latter two sciences to the problem has been mostly in the area of modelling and simulations, concentrating on collecting experimental data for the plausibility of some of the proposed theories of language evolution. As helpful as such work might be, it has to be noted with some disappointment that few of the computational approaches mentioned above have ventured deeper into isolating the importance of the still obscure issues of symbol grounding, dialogue structuring, questions, mental representations and pragmatics with a realistic set of assumptions.

The current work aims to contribute in this direction with model that is quite abstract, yet sufficiently realistic in terms of assumptions made regarding the cognitive capacities of early humans. We apply this model to a number of different hypothetical social structures among groups of agents in order to observe the communicative advantages and disadvantages of these structures with respect to supporting the emergence of natural language. First we summarize our modelling approach and then provide a technical description of the class of models explored here. We present some of our most recent experiments and discuss expected and observed outcomes, concluding with an evaluation of the results and suggestions for future work. While the results are not decisive, we hope to intrigue the reader with the overall approach and motivate our colleagues to adapt the approach to other related scenarios.

## 2   Modelling Approach

The experiments presented in this paper have been performed using the Language Evolution Workbench (LEW, [23]), which was extended by a more intuitive forgetting mechanism as well as the possibility to run simulations with different underlying social structures, as explained below. This workbench provides over 20 adjustable parameters and makes as few *a priori* assumptions as possible. The nevertheless assumed cognitive and social abilities of agents have been motivated by the more widely accepted evolutionary accounts and thus fit in well with a number of models and scenarios proposed by other authors (see Section 1 for a small selection of literature).

**Abstract Model** The LEW model is implemented at a relatively high level of abstraction, meaning that some interactions and factors of the outside world are either modelled in a very simplistic manner or not modelled at all. While we concede that such an approach might make the model open to a certain amount of criticism regarding its validity in terms of being an acceptable representation of reality, there are two arguments that should be mentioned in defence of such an approach.

First of all, a highly abstracted model of a certain system means that all the elements of such a model are distinctly observable and their effects well quantifiable. While a model with hundreds of parameters would certainly bring it closer to reality, one would find it extremely hard to distinguish between significant and insignificant parameters in such a model, as well as to observe the interactions between different parameters.

Furthermore, by starting with a simpler model, we aim to avoid the mistake of building features into it that have not yet been proven or observed well enough in other disciplines. In other terms, if one does not know the precise parameter settings for the dimensions that impinge on the problem, one should not just build in arbitrary settings as features of the model without experimenting with a range of parameter combinations first. However, since the number of such experiments grows exponentially with every free parameter in a model, we have elected to approach this issue by tackling a smaller number of features at a time, with the option of fixing the parameter values of a particular feature in case of little or no significance and moving on to the next feature, thus gradually extending the model.

Due to the abstracted nature of agent, entity and event representation, it should be noted at this point that the model is easily adjustable to represent a wide variety of social scenarios, thus making it well suited for experiments even outside the scope of language evolution. The main emphasis of the model is on observing how patterns emerge in a simulated system without there being any sort of explicit force driving the system in any particular direction.

**General Assumptions** Agents in the LEW are equipped with the ability to observe and individuate events, i.e. an abstracted sensory mechanism. Each agent individuates events according to its own perspective, as likely as not distinct from that of companions. In order to model communication, agents are assumed to be able to join in a shared attention frame around an occurring event and engage in an interaction, whereby one of the agents is assigned the intention to comment on the event and the other, listening agent understands that the observed utterance is the speaker's comment on the event and attempts to decode the meaning of the perceived symbols accordingly. These cognitive skills of attention sharing and intentionality perception have been marked as integral to the origins of language among others by [21].

Three further assumptions are relevant to symbol production and perception during interactions between agents: that agents are able to produce discernible symbols at all, that such phonemes can be combined to invent new symbols and that the transmission of symbols and phonemes occurs without noise; however, agents do not necessarily segment symbol sequences identically.[1] These assumptions are made on the grounds that language could not have possibly evolved without some sort of symbols being emitted.

The LEW fits with the so called faculty of language in the narrow sense as proposed by [12] in that the agents are equipped with the sensory, intentional and concept-mapping skills at the start, and the simulations attempt to provide an insight into how these could be combined to produce a communication system with comparable properties to a human language. Further, the LEW agents can be seen as having completed steps 1 and 2 in the accounts presented by [14] or [4], i.e. autonomously re-using and inventing new symbols from a generative unit, the phoneme.

**Social Structures and Hierarchies** The notion that social groups of one type or another play a central role within the evolution of the hominid species as such as well as the emergence of a communicative system like a proto-language in particular is apparent from a variety of evolutionary theories and modelling approaches. From the anthropological point of view, it has been repeatedly suggested that the emergence of language is strongly connected with the increase of hominid group sizes and the directly related neocortex ratio between 500,000 and 250,000 years ago (cf. [1]). Being unparalleled in any other species, this evolutionary change has become the focal point of several theories on the emergence of language. While the specific details of these theories are quite variable, two main branches can be clearly distinguished in terms of the characteristic social dynamics of the scenarios.

---

[1] While the symbols are called phonemes in the current simulations, there is no reason why these should not be representative of gestural signs. However, the physiological constraints on the production of symbols is not a part of this model.

Nearly 40 years ago, [2] postulated that the unparalleled evolutionary path of hominids is based mainly on the competition between different bands or groups of the species. While this phenomenon has been also observed in other primates to some extent, the degree of competitiveness, escalating to true warfare, is considered to be unique to the human species. In contrast to [2], [8,6] propose scenarios that are based on the evolution of *Machiavellian Intelligence* in early hominids [3]. The main difference to [2] is that the focal point in these scenarios is on group-internal organisation and cooperation, rather than inter-group competition. Accordingly, internal hierarchies play a much bigger role in these accounts, even if considered at the simplest level of having one dominant member in a group. In the current experiments, the groups in the no-male runs should roughly correspond to the competing bands in [2] and the simulations with a male – to the social structures in [8,6].

One final remark regarding our implementation of 'competition' or an 'alpha-male' is that even though it is common in social and political sciences to observe a distribution of power and influence in basically any community, our model does not involve an explicit definition of power. Consequently, we can observe the effect of being organised in a 'democratic' or an 'dictatorial' power structure, as proposed by [9], only in approximated terms of implicit influence, i.e. based on some agents' higher involvement in interactions, meaning there is never a 'semantic arbiter'.

## 3   Model Implementation

**Agents, Entities and Events** Agents in the LEW are non-physical entities (cf. [20] for embodied implementations) and are not specialized to the question of language evolution. What characterizes every agent in the LEW is solely a knowledge base and a lexicon. The knowledge base consists of all experienced events in the order in which an agent encountered these. The lexicon is represented as a set of $<Meaning, Form, Weight>$ tuples, where a *Meaning* is (a part of) an event, a *Form* is (a part of) an utterance that was either produced or heard by the agent in relation to the event, and the *Weight* is an indicator of confidence in the mapping, incremented each time it is experienced. If forgetting is enabled, these weights are then gradually decreased according to the selected forgetting function and its parameters.

*Events* are generated by selecting one of the predefined *event types*, which define the combination of arguments that is permitted for event instances of the given type (e.g. `[human, human, event]`), and filling it with acceptable arguments. An *argument* of an event can either be an entity or another event, the latter option allowing for recursive composition of events, resulting in an un-bounded meaning space based on a finite number of event types. Entities are represented as propertyless atoms, and an arbitrary number of these can be experimented with. However, in the presented work we define entities in terms of sorts, whereby two sorts are distinguished, namely `animates` and `inanimates`. Adding an abstracted layer of physical properties to entities for simulating concept formation is a possible future extension of the LEW.

**Interactions** Building on the traditions of computer simulations of language evolution, the LEW simulates interactions between agents. Every interaction in the LEW occurs between two randomly chosen agents, a speaker and a hearer, whereby an agent can also end up talking to himself if he gets picked as the hearer too (language is meant for thinking as well as communicating). The speaker is first of all presented with an event constructed as described above, e.g. `xcvww human twedf inanimate`. The speaker's task is then to individuate the event by segmenting it into meaning chunks or, in other terms, by combining parts of the event into unified segments, e.g. `[xcvww] [human] [twedf inanimate]`, which he then attempts to communicate to his conversation partner by either using an appropriate mapping from his lexicon or by inventing a new word if the meaning is new to him.

The second agent – the hearer – has the task of decoding the meaning of the heard utterance by attempting to assign (parts of) the event to (parts of) the utterance by either looking for appropriate form-meaning pairs in its lexicon or, failing to find one, by simply assuming (a part of) its own perspective on the event (e.g. `[xcvww human twedf] [inanimate]`) as the intended meaning. While this scenario presumes that both agents are knowingly communicating about the same event, their internal segmentations of the event can be, and usually are, quite different, which ensures that no omniscient meaning-form transfer occurs at any stage of the simulated interactions.

The words used by the agents in their interactions are implemented as combinations of phonemes, whereby every phoneme is represented as a pair of phones, thus mimicking the onset-nucleus structure (without the coda). When inventing a new word, the speakers use a single phoneme only. However, since agents do not possess the capacity of detecting word boundaries from an encountered utterance, hearers have the 'power' to wrongfully segment heard utterances and thus introduce larger words into their lexicons and subsequently, when acting as a speaker, into the lexicons of others.[2]

**Group Dynamics** In order to be able to perform experiments with different social structures as described in Section 2, we have extended the LEW with three parameters that determine the social organisation of a simulated population, namely the presence of a 'male' (represented as a binary variable), the number of groups $n$ ($\geq 1$) that the non-male population should be split up into and the ratios for the distribution of agents into these groups $r_1,..,r_n$, so that the size of any non-male group $C_i = r_i * (C_{total} - male)$, where $C_{total}$ is the total number of agents in the system.

---

[2] The possibility of having synchronized speech segmentation can nevertheless be explored in the LEW via the synchrony parameter, but was turned off for the presented experiments (cf. [22] for an account of experiments with synchronous transmission).

After dividing the agents into a particular social structure, we can define how they will interact with each other during the simulation using two further parameters: the male-directed communication rate $p_{male}$, defines the chances of an agent selecting the male as the hearer in an interaction, and the intra-group communication rate $p_{intra}$, which is defined as the probability of a speaker agent from group $G_i$ picking another agent from his own group (including himself) as the hearer, as opposed to an agent from groups $G_1..G_{i-1}, G_{i+1}..G_n$, after having decided that he does not wish to interact with the male. The probabilities of picking an agent from either of the neighbouring groups were distributed equally from any remaining percentage.

| Speaker \ Hearer | Male | Same group | Every other group |
|---|---|---|---|
| Male | 0 | – | $\frac{1}{n}$ |
| Non-Male | $p_{male}$ | $p_{intra}$ | $\frac{1-(p_{male}+p_{intra})}{n-1}$ |

**Table 1.** Probabilities of speaker-hearer combinations for each type of agent.

Importantly, and as can be seen in Table 1, the intra-group communication rate only applies to non-male agents, meaning that the male has equal chances of selecting any agent from any group for an interaction, except for himself. The avoidance of male self-talk is mainly motivated by the fact that a male is already involved in a much larger number of communicative bouts and may simply not have enough time to be alone and talk to himself.

## 4 Experiment Design

The goal of the presented experiments was to observe the effect of different hierarchies and social structures on the overall speed and success of communication within a group of agents. This approach extends the LEW in a way that would enable it to be used at least as a partial model for the theories of the origins of language that are based on social interactions of early humans. In particular, the experiments should provide empirical data for the possibility of language emerging in differently organized social groups, building on either the competitive account presented by [2] or the grooming theory of [8] and the corresponding comparative research by [16].

For the current experiments, all but two parameters of the LEW have been kept fixed at the following values: 9 agents divided equally into three groups of 3, with no agent addition or elimination occurring, the male-directed communication rate set to 20% in all simulations where a male was present, 100 event types with a Zipfian distribution, 41 phonemes, asynchronous utterance segmentation, frequency-based mapping retrieval, forgetting enabled and no questions.

The two varied parameters were the presence of a male in a population and the intra-group communication rate for the non-male agents whereby five different rates were experimented with – 0%, 33%, 50%, 80% and 100% – resulting in ten combinations of simulation settings that represent a variety of social structures from free circulation to full isolation.

**Expected Outcome** The goal of the experimental setup described above was to observe if a particular social structure is somehow better suited for the emergence of a group-wide communication system. The prediction that we make is that agents communicating mainly within their own group should achieve higher levels of understanding; however, these agents are expected to evolve their own sub-dialects that are quite distinct from those of other groups of the community, thus making them unable to properly cooperate with most members of the whole community, if the need for such cooperation ever occurred. The evaluation measures applied during the experiments in order to verify the postulated hypothesis and quantify the suggested effects are described in more detail below.

**Evaluation Measures** When two agents communicate with each other in the LEW, they have no access to the internal states of their interlocuting counterparts and are thus, rather disappointingly from their point of view, unable to telepathically know what the other agent is talking about. However, this does not mean that communication success is not measurable in some way. In fact, the model includes a number of various measures that allow us to observe and analyse the emerging symbol systems in a sufficiently rigorous fashion, for example by comparing the intended meaning of the speaker with the understood meaning of the hearer, either explicitly or implicitly.

From the explicit point of view, one can observe how many of the speaker's words have been actually segmented correctly, and subsequently how many of the correctly segmented words have been decoded into a meaning by the hearer that matches exactly the intended meaning of the speaker. This explicit measure can be also seen as measuring the cohesiveness of lexical overlap in two interacting agents. However, understanding can be also measured implicitly, namely without regard for the lexical items that were used to convey the meaning. So if the speaker wants to say $A$, but either by accident or because of not knowing any better, says $B$, then if the hearer, again by chance or lack of linguistic knowledge, still understands $A$ then the interaction can be seen as successful to a certain degree.

Apart from evaluating the actual communication scenarios, we also observe the lexicons of the agents to be able to draw more qualitative conclusions. For instance, the lexicon size indicates the range of expressible meanings and interpretable forms; the amount of synonymy and homonymy both inside the individual lexicons and across of the whole population tells us how similar the emergent languages are to natural languages, which seem to tolerate homonymy and avoid synonymy; while the amount of mappings shared by the whole population and the number of agents sharing a mapping on average are both good indicators of potential communicative success.

## 5   Results and Discussion

In total, 600 runs have been executed for each of the factor combinations with a total of 200 rounds of 10 interactions within each such run. The distributions of understanding success rates are presented in Figure 1(a) and suggest that there is a strong difference in the potential of language evolving in a particular group depending on the group's hierarchical and social structure.[3] In particular, one observes an increase in communicative success with the increase of intra-group communication rate ($t \geq 14.10$, $p < 0.0001$ for every level of $p_{intra}$), showing that the more isolated a group of agents is, the likelier it seems to evolve a group-internal language that serves well as a medium of communication. It should be noted that the setting with no male and a 33% intra-group communication rate corresponds to one group of 9 agents where any agent has an equal chance of being selected as either the speaker or the hearer, and is thus essentially the current baseline of the LEW as all previously conducted experiments were performed with this setup [23,22].

When considering the effect of having a male in the population, it can be noted that the addition of such a single interconnected agent is not advantageous for the whole group ($t = -12.43$, $p < 0.0001$). Most plausibly, this can be explained by the lack of opportunity for the agents to build a common language as they are in effect too occupied with attempting to communicate with the male. However, since the chances of the male being selected as the speaker are equal to those of any other agent, he does not have sufficient power to actually regulate and stabilize the language of other agents.

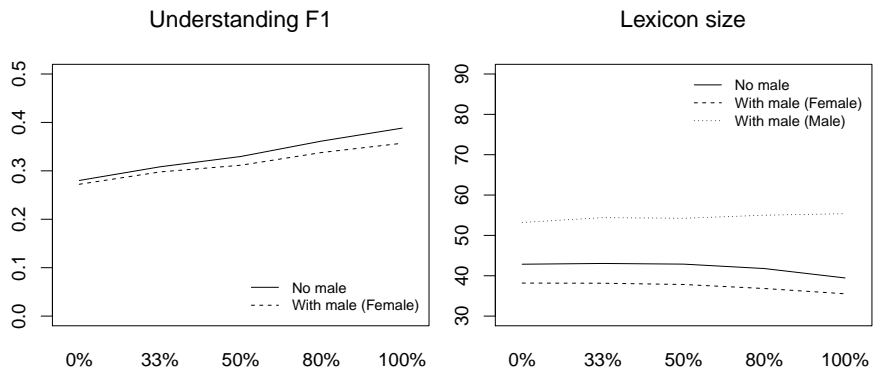

**Fig. 1.** Effect of male presence and varying intra-group communication rates (x-axis) on (a) communication success rate (F1 measure) and (b) agent lexicon size.

---

[3] Let the reader not be discouraged by the relatively low levels of understanding as it is our foremost goal to show that at least some sort of recognition of what is being talked about can be achieved at the very early stages of language evolution. In addition, [5] suggests that 'nearly' experiences can be quite stimulating even if the actual rewards are minimal.

Figure 1(b) suggests that the male lexicon in the given scenario and under the presented experimental settings tends to exhibit a larger number of both expressible meanings and interpretable forms, resulting in a bigger size overall, compared to agents from the surrounding groups ($t = 120.44$, $p < 0.0001$). This tendency is more or less a direct consequence of the setting that, regardless of the intra-group communication rate, every agent from any group interacts with the male 20% of the time. Having different members of the community constantly communicating with the male results in a deep source of linguistic data being provided to and processed by the male.

A subsequent effect of having a male in the community is that the lexicon size of an average member actually drops below those observed in communities without a unifying male-like agent ($t = -33.02$, $p < 0.0001$). The explanation is purely quantitative and essentially says that if an agent is occupied with speaking to the male 20% of the time and if the total number of interactions is kept constant, he will have less time to devote to interacting with other members of either his own or neighbouring groups of the community. This in turn reduces the amount of linguistic data that an average group-agent has a chance of observing and learning from, resulting in him learning less meaning-form mappings.

The ramifications of having a male in the presented experiments is both similar and slightly distinct from the acceleration-deceleration effect observed by [11]. On the similar side, the effects of the male acting as a hub and connecting the agents on the one hand and being overloaded with different idiolects and failing to transmit a consistent language to other agents on the other hand are certainly present in our experiments. However, it appears that another complementary effect can be observed in our case, namely that since the chances of the male acting as a speaker are equal to those of any other agent, the male appears to take a lot of linguistic data in, but not give back equally as much, thus acting more like a language server than a true hub.

Going further, a correlation between the higher intra-group communication rates and agent lexicon sizes can be observed in the figures above, most notably for the higher levels of $p_{intra}$ ($t = -17.530$, $p < 0.0001$ for $p_{intra} = 100\%$ and $t = -6.998$, $p < 0.0001$ for $p_{intra} = 80\%$; however $p = 0.337$ for $p_{intra} = 50\%$ and $p = 0.664$ for $p_{intra} = 33\%$). It is quite clearly the case that if a regular agent is sufficiently restricted to interacting with other agents of her own group then she is exposed to less linguistic variation and consequently fewer meaning-form mappings that she could possibly learn. On the other hand, the lexicon of the male increases as agents become more and more group-oriented and start developing independent dialects up to the point where they basically have their own languages ($t = 4.749$, $p < 0.0001$ for $p_{intra} = 100\%$) that are only connected by one agent – the male – who is exposed to, and basically forced into learning all three of these. For the fully isolated groups, the smaller lexicons appear to further result in a decrease of agent lexicon synonymy ($t = -11.627$, $p < 0.0001$); other settings show no such effect.

The explanation for the increase in agent-internal lexicon homonymy corresponding with the increase in the intra-group communication rate requires some clarifications as to how homonyms can emerge in a language. The most intuitive source of homonymy in the LEW is located within the utterance generation process. In particular, when an agent is presented with a meaning that he has not previously encountered and has hence no associated word for, he expresses the meaning by emitting a random string. Importantly, since we are reluctant to assume some kind of a cognitive mechanism implementing the *principle of contrast* as a given, the space of such random strings is not restricted by forms that are already present in an agent's lexicon.[4]

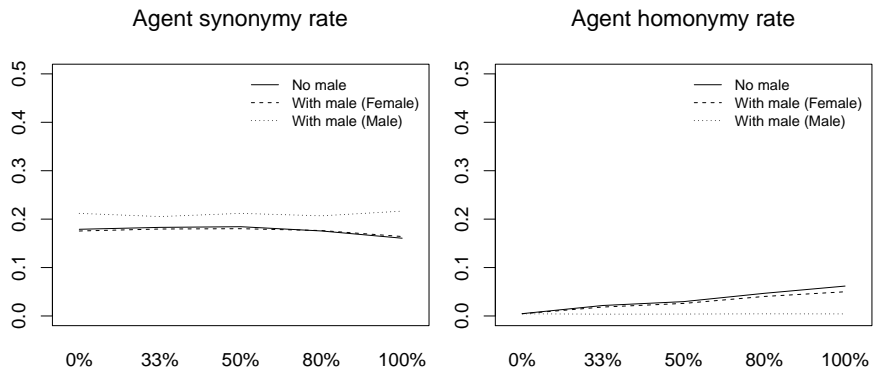

**Fig. 2.** Effect of male presence and varying intra-group communication rates (x-axis) on (a) agent lexicon synonymy and (b) agent lexicon homonymy.

Apart from the above scenario, the current implementation of the LEW also introduces a certain amount of homonymy when an agent engages in a round of self-interaction, which roughly corresponds to the primate notion of 'thinking'. What happens during self-talk is that when an agent assigns some random (even previously unused) form $f$ to a meaning $m$ that he wishes to express, he will not 'remember' that he just assigned that form to the meaning $m$ when acting as a hearer, and if he has never heard the form before, he can end up with selecting any random segment of the perceived event as the meaning, resulting in two different meanings being mapped to a single form within one interaction. This particular 'feature' of the model explains why the average homonymy levels of agent lexicons tend to increase with higher intra-group communication rates and correspondingly higher self-talk chances ($t \geq 32.78$, $p < 0.0001$ for every level of $p_{intra}$), as seen from Figure 2(b). The homonymy rates of the male lexicon are resistant to this tendency because the experiments were set up with no male self-talk, motivated in part by them being occupied with speaking to others most of the time.

The initial observations appear to suggest that on an idiolectic level, homonymy seems to creep into an agent's lexicon when the agent frequently engages in rounds of self talk and synonymy is in general governed by the level of isolation of a group from other groups that employ distinct dialects. What can not be concluded from this, however, is if these factors have an equal effect on the synonymy and homonymy ratios of the collective language of the simulated communities. It is not possible to make such conclusions because we can not see the whole picture behind the evolution of meanings and forms from the perspective of any single agent. To exemplify this, it can be imagined that the increase in agent lexicon synonymy does indeed signify an introduction of a number of new and redundant words to the language. However, it can also be the case that the agents within every group previously associated an immensely large amount of words with a few especially frequent meanings, resulting in low lexicon synonymy, but did not know any of the mappings common to the dialects employed by the agents of neighbouring groups, thus making the language of the community possess an extremely high level of overall synonymy. However, if each of these agents were to learn one form for every meaning from each of the other dialects and at the same time forget the redundant synonyms within his own group dialect, global synonymy would actually decrease, courtesy of these forgotten forms, yet the agent lexicon synonymy would increase dramatically.

---

[4] The exact chance of a previously used form being assigned to a new meaning at time $t$ equals the number of forms known by the agent at $t$ divided by the space of possible combinations of two phonemes.

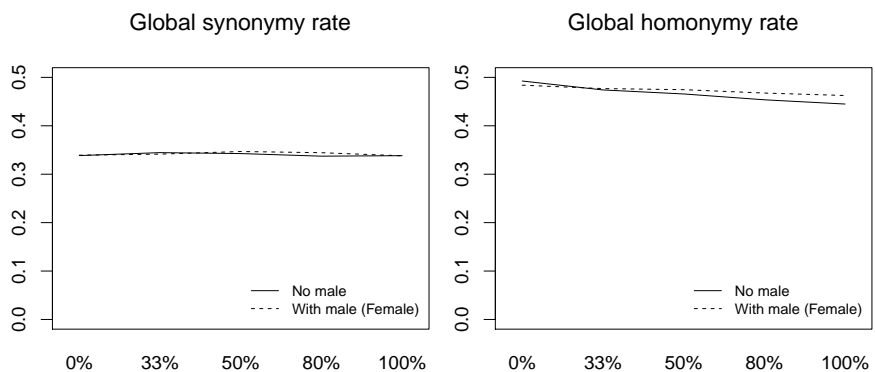

**Fig. 3.** Effect of male presence and varying intra-group communication rates (x-axis) on (a) global lexicon synonymy and (b) global lexicon homonymy.

This last scenario of increasing idiolectic levels and decreasing global levels does indeed apply to homonymy in the emergent language of the presented simulations (cf. Figures 2(b) and 3(b)). Accordingly, the global homonymy levels are slightly dropping with the increase of intra-group communication ($t \leq -6.703$, $p < 0.0001$ for every level of $p_{intra}$), despite the opposite tendency being observed for the individual lexicon homonymy. On the other hand, the effect of group isolation on global synonymy levels appears to be well balanced between lower dialectic and idiolectic synonymy (cf. Figure 2(a)) and the growing idiosyncrasy of the dialects, resulting in an insignificant overall effect ($p > 0.1$ for every level of $p_{intra}$), as exhibited on Figure 3(a).

## 6   Conclusions

To summarize the results, three main conclusions can be drawn from the simulation runs described above. The first is that smaller and more isolated groups have a clear advantage over larger or more actively interacting groups in terms of evolving a reliable and useful communication system. The second is that having a single interconnected agent who is asymmetrically favoured as a hearer, and equally likely to be a speaker has no positive influence on the disjoint groups. The third observation is that while the communicative success of agents' interactions in the closely connected groups is inferior to that of the isolated groups: the more 'social' agents appear to learn a significantly larger number of mappings without a commensurate increase in lexical synonymy, resulting in higher chances of understanding utterances by a wider range of agents, including those with different dialects.

Going back to the two general social theories of language evolution presented in Section 2, our results do not truly give support to either of these. In particular, the conclusion that isolated groups tend to outperform more 'social' groups is seemingly at odds with the notion that language evolution was driven by either inter- or intra-group cooperation which lies at the heart of both introduced scenarios. In addition, the theories that are based on group-internal cooperation and should thus involve some sort of hierarchy like having a strongly interconnected alpha-male member are not reinforced by our current simulations either, although we are willing to concede that these structures are still extremely rudimentary and require further investigation.

However, when comparing our results with those of other computational modellers, it appears that the simulations in which agents interact with a higher number of other agents coincide with the results presented by [17,24], both of which also eschew telepathic meaning transfer and explicit feedback. The experiments conducted by these authors have exhibited similarly high levels of synonymy on both the agent-internal and language-global levels, which is not characteristic of human languages that we know of: all extant models are thus missing some biological or cognitive mechanism that would explain synonymy avoidance as an emergent property, and not a consequence of a pre-programmed principle of discriminative reasoning.

Either way the onus remains on the modelling approaches to show that not only can the presented models attest for the emergence of a communicative system, but equally that this system does indeed significantly resemble language as we know it, which so far is yet to be the case. In particular, models of language evolution that implement such aiding mechanisms as e.g. explicit feedback converge on a perfect communication system without synonyms or homonyms, and without any additional variation after the convergence has been reached [20]. While such a system might seem quite optimal in terms of communicative efficiency, it certainly does not resemble a human language.

The shortcomings of such models have been pointed out by several authors, including [22,17], with [18] attempting to solve the problem by building a model without omniscience or explicit feedback, but with a number of representational, interpretational and social constraints instead, e.g. the "whole object" bias and the principle of contrast. The experimental results of this model suggest that integrating such constraints does indeed improve the communicative success of the emergent language, while keeping it comparable to existing human languages. However, embodiedment of these additional constraints is yet to be conclusively proven to exist in primates or even humans; work by [10,7] suggests that these constraints are perhaps not present in our brains at all, or at least not in the strong version as implemented in [18].

In conclusion, the experiments presented in this paper hopefully provide an outline of what kind of simulations of group dynamics are possible with the help of the LEW, along with its promise and shortcomings. While the current results are not yet comprehensive enough to speak for themselves in their generality or importance, we hope to have made a case for the experimental approach *per se*. We see this as representative of our programme of experiments on the effects of social structures and group dynamics in language evolution. Future work would certainly benefit from a further extension of the LEW to allow for more complex group dynamics settings, including multi-level hierarchies and coalitions between selected groups.